\documentclass{article} % For LaTeX2e
\usepackage{iclr2024_conference,times}

\usepackage{amsmath,amsfonts,bm}

\def\eqref#1{equation~\ref{#1}}
\def\1{\bm{1}}

\def\vtheta{{\bm{\theta}}}

\def\vc{{\bm{c}}}

\def\vf{{\bm{f}}}

\def\vx{{\bm{x}}}

\def\vz{{\bm{z}}}

\DeclareMathAlphabet{\mathsfit}{\encodingdefault}{\sfdefault}{m}{sl}
\SetMathAlphabet{\mathsfit}{bold}{\encodingdefault}{\sfdefault}{bx}{n}

\usepackage{hyperref}
\usepackage{url}

\usepackage{graphicx}
\usepackage{subfigure}
\usepackage{float}
\usepackage{tipa}

\usepackage{bm}

\usepackage{multirow}
\usepackage{amsthm}
\usepackage{wrapfig}
\usepackage{multicol}
\usepackage{algorithm}
\usepackage{algorithmic}
\usepackage{amsmath}
\usepackage{makecell}
\usepackage[textsize=tiny]{todonotes}
\usepackage{booktabs}
\usepackage{amssymb}

\usepackage{enumitem}

\title{LCM-LoRA: A Universal Stable-Diffusion \\ Acceleration Module}

\author{Simian Luo$^{*,1}$ \quad Yiqin Tan\thanks{Leading Authors} \ $^{,1}$ \quad
Suraj Patil$^{\dagger,2}$ \quad Daniel Gu\thanks{Core Contributors} \  \quad Patrick von Platen$^{2}$ \\ \textbf{Apolinário Passos$^{2}$ \quad Longbo Huang$^{1}$ \quad Jian Li$^{1}$ \quad Hang Zhao$^{1}$} \\
$^1$ IIIS, Tsinghua University \quad $^2$ Hugging Face \\ 
\texttt{\{luosm22, tyq22\}@mails.tsinghua.edu.cn} \\
\texttt{\{suraj, patrick, apolinario\}@huggingface.co} \\
\texttt{\{dgu8957\}@gmail.com}\\
\texttt{\{longbohuang, lijian83, hangzhao\}@tsinghua.edu.cn}
}

\def\@onedot{\ifx\@let@token.\else.\null\fi\xspace}

\makeatother

\iclrfinalcopy % Uncomment for camera-ready version, but NOT for submission.

\begin{document}

\maketitle

\begin{abstract}

Latent Consistency Models \textbf{(LCMs)} \citep{luo2023latent} have achieved impressive performance in accelerating text-to-image generative tasks, producing high-quality images with minimal inference steps. LCMs are distilled from pre-trained latent diffusion models (LDMs), requiring only $\sim$32 A100 GPU training hours. This report further extends LCMs' potential in two aspects: First, by applying LoRA distillation to Stable-Diffusion models including SD-V1.5 \citep{rombach2022high}, SSD-1B \citep{segmind2023ssd1b}, and SDXL \citep{podell2023sdxl}, we have expanded LCM's scope to larger models with significantly less memory consumption, achieving superior image generation quality. Second, we identify the LoRA parameters obtained through LCM distillation as a \textit{universal Stable-Diffusion acceleration module}, named \textbf{LCM-LoRA}. LCM-LoRA can be directly plugged into various Stable-Diffusion fine-tuned models or LoRAs \textbf{without training}, thus representing a universally applicable accelerator for diverse image generation tasks. Compared with previous numerical PF-ODE solvers such as DDIM \citep{song2020denoising}, DPM-Solver \citep{lu2022dpm,lu2022dpm++}, LCM-LoRA can be viewed as a {plug-in neural PF-ODE solver} that possesses strong generalization abilities. Project page: \url{https://github.com/luosiallen/latent-consistency-model}.
\end{abstract}

\begin{figure}[h] 
\begin{centering}
\includegraphics[scale=0.37]{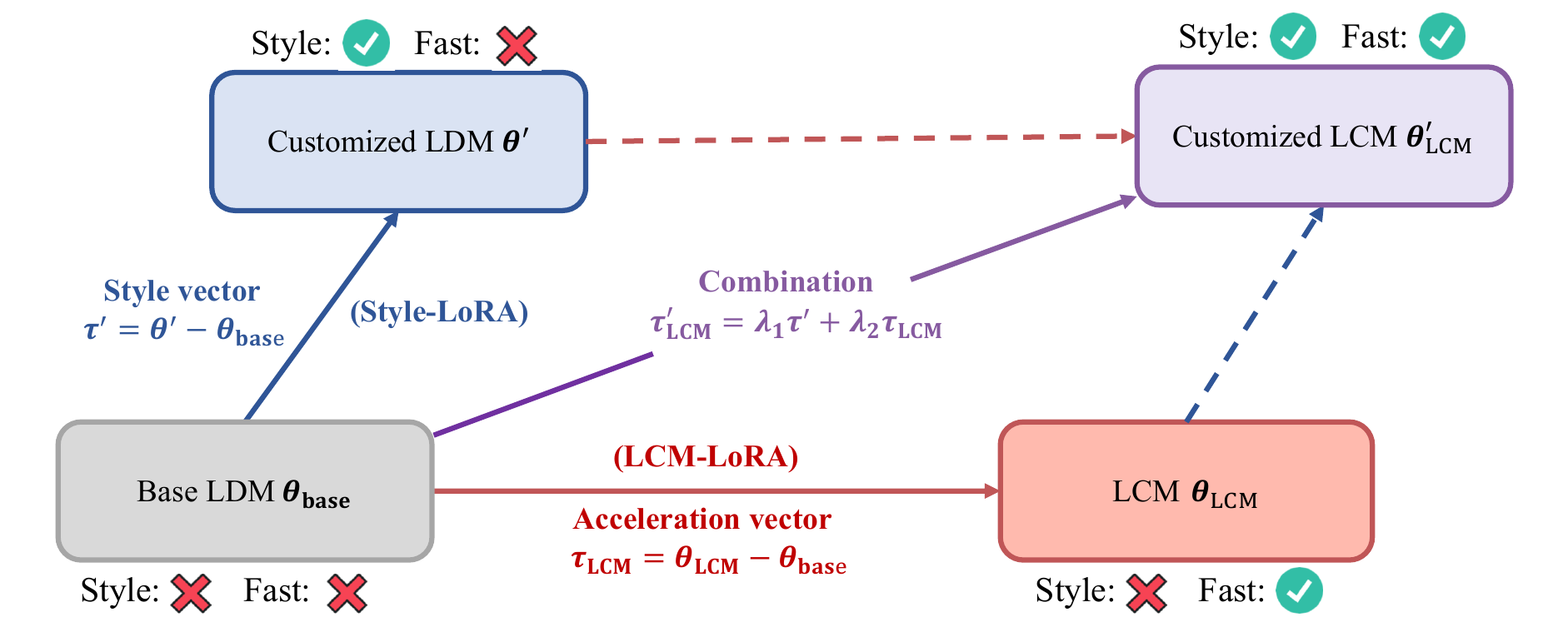}
 \vspace{-0.1in}
\caption{\footnotesize Overview of LCM-LoRA. By introducing LoRA into the distillation process of LCM, we significantly reduce the memory overhead of distillation, which allows us to train larger models, e.g., SDXL and SSD-1B, with limited resources. More importantly, LoRA parameters obtained through LCM-LoRA training (`acceleration vector') can be directly combined with other LoRA parameters (`style vetcor') obtained by fine-tuning on a particular style dataset. Without any training, the model obtained by a linear combination of the acceleration vector and style vetcor acquires the ability to generate images of a specific painting style in minimal sampling steps.}\label{fig:lcm_lora}
\end{centering}
\vspace{-0.1in}
\end{figure}

\section{Introduction}
% Latent Diffusion Models (LDMs) \citep{rombach2022high} have achieved great success in synthesizing highly realistic and artistic images given different types of conditions (e.g. text prompts, sketch, layout). However, LDMs suffer from the slow iterative reverse sampling process to generate data, which further restricts their real-time applicability and results in an inferior user experience. Although recent research has been attempting to accelerate diffusion models, there is currently no superior open-source model or acceleration method that can truly achieve real-time generation on consumer-grade GPUs. Two kinds of methods have been adopted to accelerate LDMs. One is to use more advanced ODE-Solvers, like DDIM \citep{song2020denoising}, DPM-Solver \citep{lu2022dpm}, DPM-Solver++ \citep{lu2022dpm++}, to speed up LDMs' generation. The other one is to use distill LDMs \hang{citations}. The former still requires $\geq 10$ inference steps to obtain satisfying results and is also restricted by the extra computational cost of classifier-free guidance \citep{ho2022classifier}, while the latter like Guided-Distill \citep{meng2023distillation} is limited in practicality due to its substantial computational costs. 

Latent Diffusion Models (LDMs) \citep{rombach2022high} have been pivotal in generating highly detailed and creative imagery from various inputs such as text and sketches. Despite their success, the slow reverse sampling process inherent to LDMs hampers real-time application, compromising the user experience. Current open-source models and acceleration techniques have yet to bridge the gap to real-time generation on standard consumer GPUs. Efforts to accelerate LDMs generally fall into two categories: the first involves advanced ODE-Solvers, like DDIM \citep{song2020denoising}, DPM-Solver \citep{lu2022dpm} and  DPM-Solver++ \citep{lu2022dpm++}, to expedite the generation process. The second strategy involves distillation of LDMs to streamline their functioning. The ODE-Solver methods, despite reducing the number of inference steps needed, still demand a significant computational overhead, especially when incorporating classifier-free guidance \citep{ho2022classifier}. Meanwhile, distillation methods such as Guided-Distill \citep{meng2023distillation}, although promising, face practical limitations due to their intensive computational requirements. The quest for a balance between speed and quality in LDM-generated imagery continues to be a challenge in the field.

% Recently, inspired by Consistency Models (CMs) \citep{song2023consistency}, Latent Consistency Models \textbf{(LCMs)} \citep{luo2023latent} are proposed to solve the slow sampling problem by viewing the reverse guided diffusion process as solving an augmented probability flow ODE (PF-ODE). LCMs are designed to directly predict the solution of such PF-ODE in the latent space, instead of using the numerical ODE-Solver to iteratively solve it, allowing very few inference steps ($1\sim4$) to synthesize high-resolution images with great quality. Furthermore, LCMs are highly efficient in terms of distillation methods, requiring only approximately 32 A100 training hours for $2\sim4$-step inference. 

% Building upon LCMs, Latent consistency finetuning (LCF) \citep{luo2023latent} is a fine-tuning method that enables direct fine-tuning a pretrained LCM rather than a teacher diffusion model. However, to enable fast inference for customized datasets, such as anime, photo-realistic, and fantasy imagery, additional training steps are required: either using Latent Consistency Distillation (LCD) \citep{luo2023latent} to distill a pre-trained LDM into an LCM, or directly fine-tuning an LCM with Latent Consistency Finetuning (LCF) \citep{luo2023latent}. The additional training time hinders LCMs' scalability. This raises an important question: \textit{Is it possible to achieve fast inference on customized datasets without additional training?}

Recently, Latent Consistency Models (LCMs) \citep{luo2023latent} have emerged, inspired by Consistency Models (CMs) \citep{song2023consistency}, as a solution to the slow sampling issue in image generation. LCMs approach the reverse diffusion process by treating it as an augmented probability flow ODE (PF-ODE) problem. They innovatively predict the solution in the latent space, bypassing the need for iterative solutions through numerical ODE-Solvers. This results in a remarkably efficient synthesis of high-resolution images, taking only 1 to 4 inference steps. Additionally, LCMs stand out in terms of distillation efficiency, requiring merely 32 A100 training hours for a minimal-step inference.

Building on this, Latent Consistency Finetuning (LCF) \citep{luo2023latent} has been developed as a method to fine-tune pre-trained LCMs without starting from the teacher diffusion model. For specialized datasets—like those for anime, photo-realistic, or fantasy images—additional steps are necessary, such as employing Latent Consistency Distillation (LCD) \citep{luo2023latent} to distill a pre-trained LDM into an LCM or directly fine-tuning an LCM using LCF. However, this extra training can be a barrier to the quick deployment of LCMs across diverse datasets, posing the critical question of whether fast, training-free inference on custom datasets is attainable.

To answer the above question, we introduce \textbf{LCM-LoRA}, a \textit{universal training-free acceleration module} that can be directly plugged into various Stable-Diffusion (SD) \citep{rombach2022high} fine-tuned models or SD LoRAs \citep{hu2021lora} to support fast inference with minimal steps. Compared to earlier numerical probability flow ODE (PF-ODE) solvers such as DDIM \citep{song2020denoising}, DPM-Solver \citep{lu2022dpm}, and DPM-Solver++ \citep{lu2022dpm++}, LCM-LoRA represents a novel class of neural network-based PF-ODE solvers module. It demonstrates robust generalization capabilities across various fine-tuned SD models and LoRAs.

% \begin{figure}[t] 
% \begin{centering}
% \includegraphics[scale=0.37]{Figure/lora_teaser_v2.pdf}
%  \vspace{-0.1in}
% \caption{Overview of LCM-LoRA. By introducing LoRA into the distillation process of LCM, we can significantly reduce the explicit memory overhead of distilling LCM, which allows us to train larger models, e.g., SDXL and SSD-1B, with limited resources.In addition, LoRA parameters obtained through LCM-LoRA training (`accelerating vector') can be directly combined with other LoRA parameters (`style vetcor') obtained by fine-tuning on a particular style dataset. Without any additional training, the model obtained by linear combination of the accelerating vector and style vetcor has the ability to obtain images of a specific painting style in minimal sampling steps.}\label{fig:lcm_lora}
% \end{centering}
% \vspace{-0.1in}
% \end{figure}

\section{Related Work}

\paragraph{Consistency Models} \cite{song2023consistency} have showcased the remarkable potential of consistency models (CMs), a novel class of generative models that enhance sampling efficiency without sacrificing the quality of the output. These models employ a consistency mapping technique that deftly maps points along the Ordinary Differential Equation (ODE) trajectory to their origins, thus enabling expeditious one-step generation. Their research specifically targets image generation tasks on ImageNet 64x64 \citep{deng2009imagenet} and LSUN 256x256 \citep{yu2015lsun}, demonstrating CMs' effectiveness in these domains. Further advancing the field, \cite{luo2023latent} has pioneered latent consistency models \textbf{(LCMs)} within the text-to-image synthesis landscape. By viewing the guided reverse diffusion process as the resolution of an augmented Probability Flow ODE (PF-ODE), LCMs adeptly predict the solution of such ODEs in latent space. This innovative approach significantly reduces the need for iterative steps, thereby enabling the rapid generation of high-fidelity images from text inputs and setting a new standard for state-of-the-art performance on LAION-5B-Aesthetics dataset \citep{schuhmann2022laion}.

\paragraph{Parameter-Efficient Fine-Tuning} Parameter-Efficient Fine-Tuning (PEFT) \citep{houlsby2019parameter} enables the customization of pre-existing models for particular tasks while limiting the number of parameters that need retraining. This reduces both computational load and storage demands. Among the assorted techniques under the PEFT umbrella, Low-Rank Adaptation (LoRA) \citep{hu2021lora} stands out. LoRA's strategy involves training a minimal set of parameters through the integration of low-rank matrices, which succinctly represent the required adjustments in the model's weights for fine-tuning. In practice, this means that during task-specific optimization, only these matrices are learned and the bulk of pre-trained weights are left unchanged. Consequently, LoRA significantly trims the volume of parameters to be modified, thereby enhancing computational efficiency and permitting model refinement with considerably less data.

\paragraph{Task Arithmetic in Pretrained Models} Task arithmetic \citep{ilharco2022editing, ortiz2023task, zhang2023composing} has become a notable method for enhancing the abilities of pre-trained models, offering a cost-effective and scalable strategy for direct edits in weight space. By applying fine-tuned weights of different tasks to a model, researchers can improve its performance on these tasks or induce forgetting by negating them.  Despite its promise, the understanding of task arithmetic's full potential and the principles that underlie it remain areas of active exploration.

\section{LCM-LoRA}

\subsection{LoRA Distillation for LCM}\label{subsec:lora-xl}

The Latent Consistency Model (LCM) \citep{luo2023latent} is trained using a one-stage guided distillation method, leveraging a pre-trained auto-encoder's latent space to distill a guided diffusion model into an LCM. This process involves solving an augmented Probability Flow ODE (PF-ODE), a mathematical formulation that ensures the generated samples follow a trajectory that results in high-quality images. The distillation focuses on maintaining the fidelity of these trajectories while significantly reducing the number of required sampling steps. The method includes innovations like the Skipping-Steps technique to quicken convergence. The pseudo-code of LCD is provided in Algorithm~\ref{alg:LCD}.

\begin{algorithm}[H]
    \begin{minipage}{\linewidth}
    \begin{footnotesize}
        \caption{Latent Consistency Distillation (LCD) \citep{luo2023latent}}\label{alg:LCD}
        \begin{algorithmic}
            \STATE \textbf{Input:} dataset $\mathcal{D}$, initial model parameter $\vtheta$, learning rate $\eta$, {ODE solver $\Psi(\cdot,\cdot,\cdot, \cdot)$}, distance metric $d(\cdot,\cdot)$, EMA rate $\mu$, {noise schedule $\alpha(t),\sigma(t)$, guidance scale $[w_{\text{min}},w_{\text{max}}]$, skipping interval $k$, and encoder $E(\cdot)$}
            \STATE {Encoding training data into latent space: $\mathcal{D}_z=\{(\vz,\vc)|\vz=E(\vx),(\vx,\vc)\in\mathcal{D}\}$}
            \STATE $\vtheta^-\leftarrow\vtheta$
            \REPEAT
            \STATE Sample $(\vz,\vc) \sim \mathcal{D}_z$, {$n\sim \mathcal{U}[1,N-k]$ and $\omega\sim [\omega_\text{min},\omega_\text{max}]$}
            \STATE Sample $\vz_{t_{n+k}}\sim \mathcal{N}(\alpha(t_{n+k})\vz;\sigma^2(t_{n+k})\mathbf{I})$
            \STATE $\begin{aligned}{\hat{\vz}^{\Psi,\omega}_{t_n}\leftarrow \vz_{t_{n+k}}+(1+\omega)\Psi(\vz_{t_{n+k}},t_{n+k},t_n,\vc)-\omega\Psi(\vz_{t_{n+k}},t_{n+k},t_n,\varnothing)}\end{aligned}$
            \STATE $\begin{aligned}\mathcal{L}(\vtheta,\vtheta^-; \Psi)\leftarrow {d(\vf_\vtheta(\vz_{t_{n+k}},\omega,\vc, t_{n+k}),\vf_{\vtheta^-}(\hat{\vz}^{\Psi,\omega}_{t_n},\omega,\vc, t_n))}\end{aligned}$
          \STATE $\vtheta\leftarrow\vtheta-\eta\nabla_\vtheta\mathcal{L}(\vtheta,\vtheta^-)$
            \STATE $\vtheta^-\leftarrow \text{stopgrad}(\mu\vtheta^-+(1-\mu)\vtheta)$
            \UNTIL convergence
        \end{algorithmic} 
        \end{footnotesize}
    \end{minipage}
\end{algorithm}

Since the distillation process of Latent Consistency Models (LCM) is carried out on top of the parameters from a pre-trained diffusion model, we can consider latent consistency distillation as a fine-tuning process for the diffusion model. This allows us to employ parameter-efficient fine-tuning methods, such as LoRA (Low-Rank Adaptation) \citep{hu2021lora}. LoRA updates a pre-trained weight matrix by applying a low-rank decomposition. Given a weight matrix $W_0 \in \mathbb{R}^{d \times k} $, the update is expressed as $ W_0 + \Delta W = W_0 + BA $, where $ B \in \mathbb{R}^{d \times r} $, $ A \in \mathbb{R}^{r \times k} $, and the rank $ r \leq \min(d, k) $. During training, $ W_0 $ is kept constant, and gradient updates are applied only to $ A $ and $ B $. The modified forward pass for an input $ x $ is:
\begin{equation}
    h = W_0x + \Delta W x = W_0x + BAx.
\end{equation}
In this equation, $ h $ represents the output vector, and the outputs of $ W_0 $ and $ \Delta W = BA $ are added together after being multiplied by the input $ x $. By decomposing the full parameter matrix into the product of two low-rank matrices, LoRA significantly reduces the number of trainable parameters, thereby lowering memory usage.
Table~\ref{tb:lora_param_num} compares the total number of parameters in the full model with the trainable parameters when using the LoRA technique. It is evident that by incorporating the LoRA technique during the LCM distillation process, the quantity of trainable parameters is significantly reduced, effectively decreasing the memory requirements for training.

\begin{table}[H]
\label{tb:lora_param_num}
\centering
\begin{tabular}{l|ccc}
\toprule
Model & SD-V1.5 & SSD-1B & SDXL  \\
\midrule
\# Full Parameters & 0.98B & 1.3B & 3.5B \\
\# LoRA Trainable Parameters & 67.5M  & 105M  & 197M  \\
\bottomrule
\end{tabular}
\vspace{-0.1in}
\caption{Full parameter number and trainable parameter number with LoRA for SD-V1.5 \citep{rombach2022high}, SSD-1B \citep{segmind2023ssd1b} and SDXL \citep{podell2023sdxl}.}
\end{table}

\begin{figure}[t] 
\begin{centering}
\includegraphics[scale=0.2]{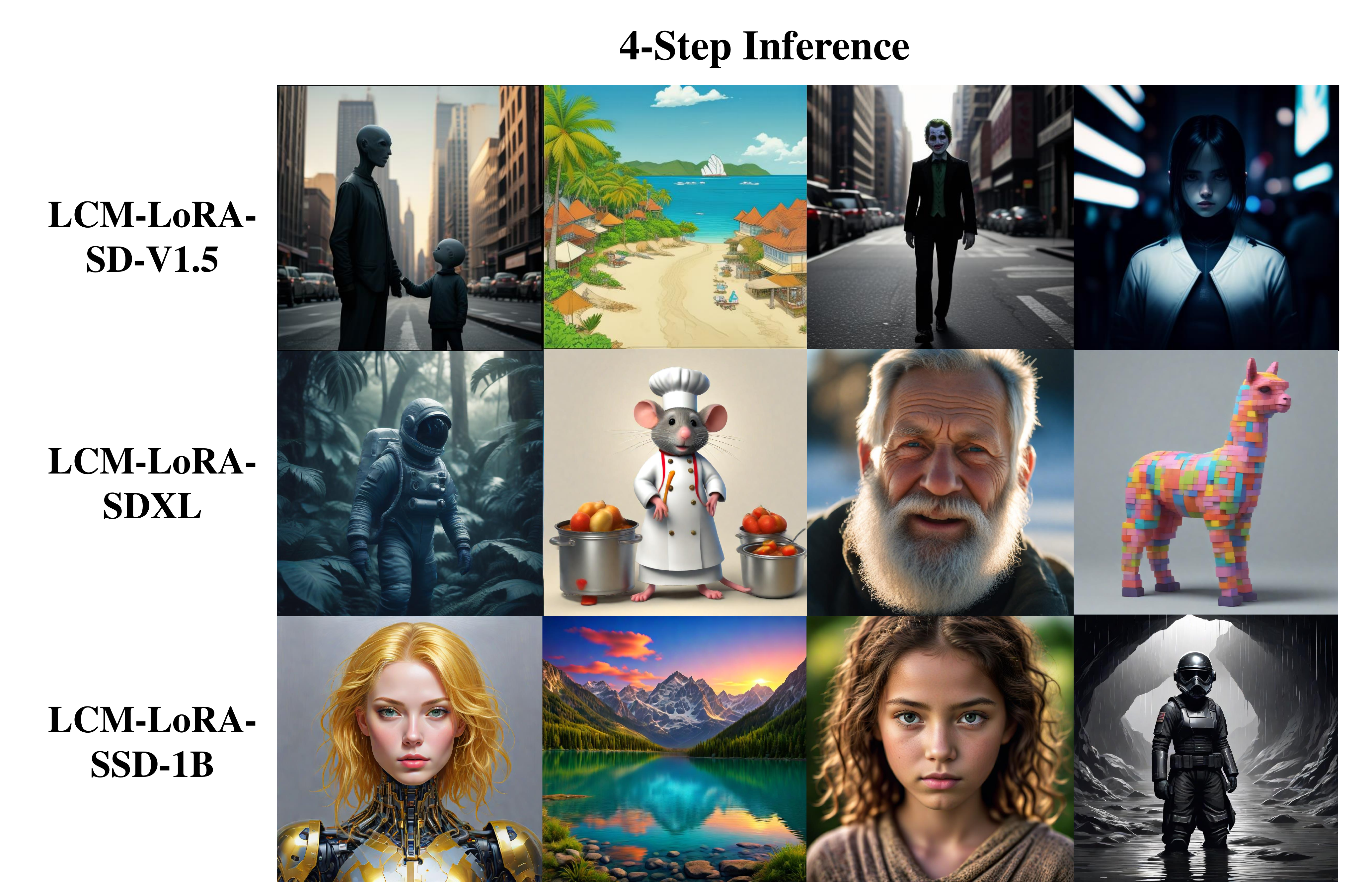}
\vspace{-0.2in}
\caption{\footnotesize{Images generated using latent consistency models distilled from different pretrained diffusion models. We generate 512$\times$512 resolution images with LCM-LoRA-SD-V1.5 and 1024$\times$1024 resolution images with LCM-LoRA-SDXL and LCM-LoRA-SSD-1B. We use a fixed classifier-free guidance scale $\omega=7.5$ for all models during the distillation process. All images were obtained by 4-step sampling .\label{fig:result1}}}
\end{centering}
\end{figure}

\cite{luo2023latent} primarily distilled the base stable diffusion model, such as SD-V1.5 and SD-V2.1. We extended this distillation process to more powerful models with enhanced text-to-image capabilities and larger parameter counts, including SDXL \citep{podell2023sdxl} and SSD-1B \citep{segmind2023ssd1b}. Our experiments demonstrate that the LCD paradigm adapts well to larger models. The generated results of different models are displayed in Figure~\ref{fig:result1}.

\subsection{LCM-LoRA as Universal Acceleratiion Module}
\begin{figure}[t] 
\begin{centering}
\includegraphics[scale=0.54]{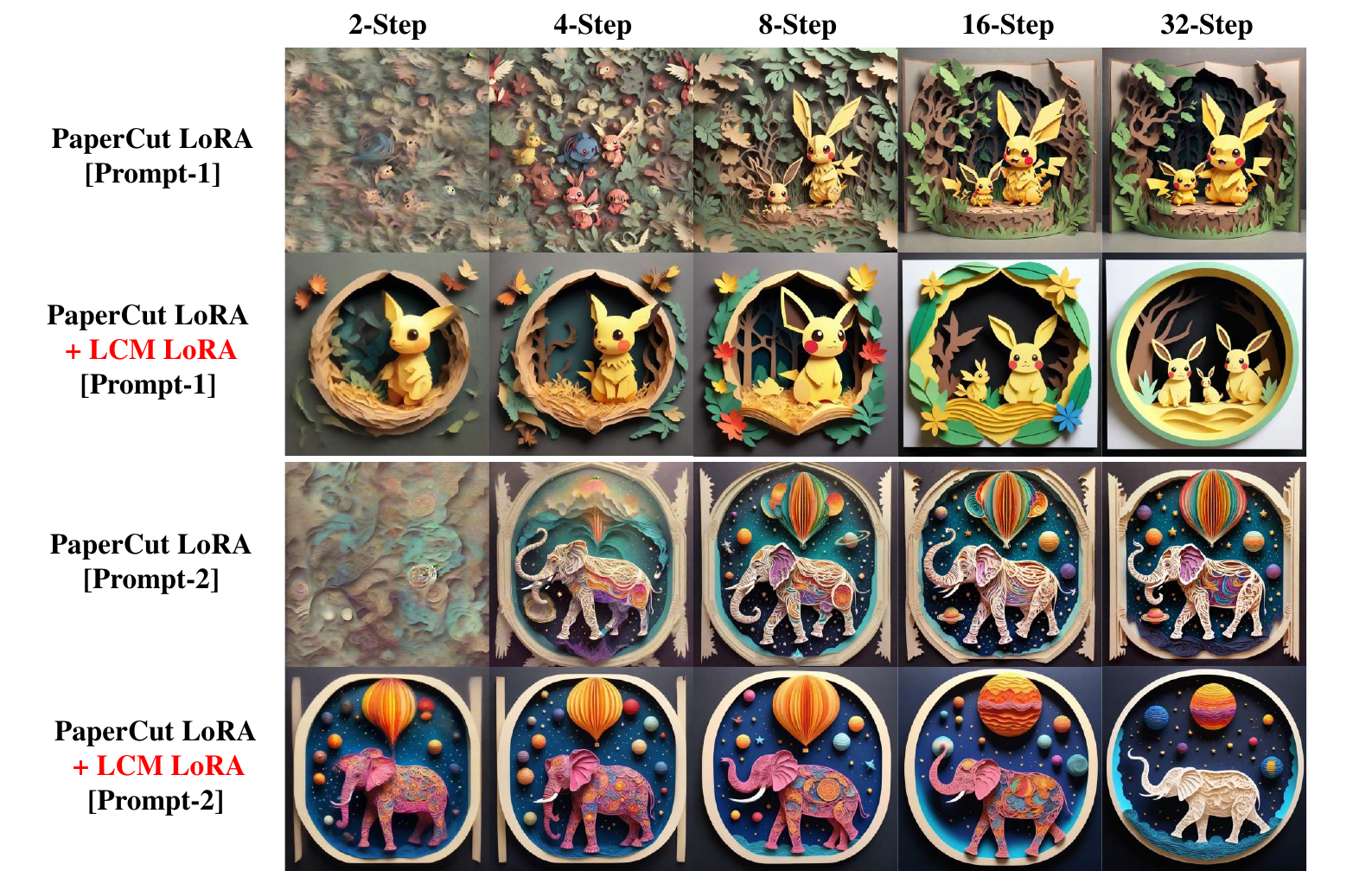}
\vspace{-0.2in}
\caption{\footnotesize{The generation results of the specific style LoRA parameters and the combination with LCM-LoRA parameters. We use SDXL as the base model. All images are 1024$\times$1024 resolution. We select LoRA parameters fine-tuned on specific painting style datasets and combine them with LCM-LoRA parameters. We compare the quality of images generated by these models at different sampling steps. For the original LoRA parameters, we use DPM-Solver++ \citep{lu2022dpm++} sampler and classifier-free guidance scale $\omega=7.5$. For the parameters obtained after combining LCM-LoRA with specific style LoRA, we use LCM's multi-step sampler. We use $\lambda_1=0.8$ and $\lambda_2=1.0$ for the combination.}\label{fig:result2}}
\end{centering}
\end{figure}

Based on parameter-efficient fine-tuning techniques, such as LoRA, one can fine-tune pretrained models with substantially reduced memory requirements. Within the framework of LoRA, the resultant LoRA parameters can be seamlessly integrated into the original model parameters. In Section~\ref{subsec:lora-xl}, we demonstrate the feasibility of employing LoRA for the distillation process of Latent Consistency Models (LCMs). On the other hand, one can fine-tune on customized datasets for specific task-oriented applications. There is now a broad array of fine-tuning parameters available for selection and utilization. We discover that the LCM-LoRA parameters can be directly combined with other LoRA parameters fine-tuned on datasets of particular styles. Such an amalgamation yields a model capable of generating images in specific styles with minimal sampling steps, without the need for any further training. As shown in Figure~\ref{fig:lcm_lora}, denote the LCM-LoRA fine-tuned parameters as $\boldsymbol{\tau}_{\text{LCM}}$, which is identified as the ``acceleration vector", and the LoRA parameters fine-tuned on customized dataset as $\boldsymbol{\tau}'$, which is the ``style vector", we find that an LCM which generates customized images can be obtained as  
\begin{equation}
\vtheta_{\text{LCM}}'=\vtheta_{\text{pre}}+\boldsymbol{\tau}'_{\text{LCM}},    
\end{equation}
where 
\begin{equation}
\boldsymbol{\tau}'_{\text{LCM}}=\lambda_1\boldsymbol{\tau}'+\lambda_2\boldsymbol{\tau}_{\text{LCM}}    
\end{equation}
is the linear combination of acceleration vector $\boldsymbol{\tau}_{\text{LCM}}$ and style vector $\boldsymbol{\tau}'$. Here $\lambda_1$ and $\lambda_2$ are hyperparameters. The generation results of the specific style LoRA parameters and their combination with LCM-LoRA parameters are shown in Figure~\ref{fig:result2}. Note that we do not make further training on the combined parameters.

\section{Conclusion}
We present LCM-LoRA, a universal training-free acceleration module for Stable-Diffusion (SD). LCM-LoRA can serve as an independent and efficient neural network-based solver module to predict the solution of PF-ODE, enabling fast inference with minimal steps on various finetuned SD models and SD LoRAs. Extensive experiments on text-to-image generation have demonstrated LCM-LoRA's strong generalization capabilities and superiority.

\section{Contribution \& Acknowledgement}
This work builds upon Simian Luo and Yiqin Tan's Latent Consistency Models (LCMs) \citep{luo2023latent}. Based on LCMs, Simian Luo wrote the original LCM-SDXL distillation code, and together with Yiqin Tan, primarily completed this technical report. Yiqin Tan discovered the arithmetic property of LCM parameters. Suraj Patil first completed the training of LCM-LoRA, discovering its strong generalization abilities, and conducted most of the training. Suraj Patil and Daniel Gu conducted excellent refactoring of the original LCM-SDXL codebase and improved training efficiency, seamlessly integrating it into the Diffusers library. Patrick von Platen revised and polished this technical report, as well as integrating LCM into the Diffusers library. Longbo Huang, Jian Li, Hang Zhao co-advised the original LCMs paper, and polished this technical report. We further thanks Apolinário Passos and Patrick von Platen for making excellent LCMs demo and deployment. 
We also want to thank Sayak Paul and Pedro Cuenca for helping with writing documentation as well as Radamés Ajna for creating demos.
We appreciate the computing resources provided by the Hugging Face Diffusers teams to support our experiments. Finally, we value the insightful discussions from LCM community members.

\bibliography{iclr2024_conference}
\bibliographystyle{iclr2024_conference}

\appendix

\end{document}